\documentclass[]{myifcolog}
\usepackage{lineno} 
\usepackage{hyperref}
\modulolinenumbers[5]
\usepackage[utf8]{inputenc}
\usepackage{graphicx}
\usepackage{multirow}
\usepackage{amsmath}
\usepackage{amsmath}
\usepackage{subfig}
\usepackage{url}

\DeclareMathSymbol{\mlq}{\mathord}{operators}{``}
\DeclareMathSymbol{\mrq}{\mathord}{operators}{`'}

\begin{document}
\title{Modelling Identity Rules with Neural Networks}

\titlerunning{Modelling Identity Rules with Neural Networks}

\addauthor[t.e.weyde@city.ac.uk]{Tillman Weyde}{Research Centre for Machine Learning, Department of Computer Science, City, University of London, United Kingdom}
\addauthor[radha.kopparti@city.ac.uk]{Radha Manisha Kopparti\thanks{Funded by a PhD studentship from City, University of London}}{Research Centre for Machine Learning, Department of Computer Science, City, University of London, United Kingdom}
\authorrunning{Weyde and Kopparti}

\titlethanks{We would like to thank the anonymous reviewers of this article for their valuable comments and suggestions that helped to improve this article.}

\maketitle

\begin{abstract}
In this paper, we show that standard feed-forward and recurrent neural networks fail to learn abstract patterns based on identity rules. 
We propose 
Relation Based Pattern (RBP) extensions to neural network structures that solve this problem and answer, as well as raise, questions about integrating structures for inductive bias into neural networks.

Examples of abstract patterns are the sequence patterns ABA and ABB where A or B can be any object. 
These  were introduced by  Marcus et al (1999) who also found that 7 month old infants recognise these patterns in sequences that  use an unfamiliar vocabulary while simple recurrent neural networks do not.
This result has been contested in the literature but it is confirmed by our experiments. 
We also show that the inability to generalise extends to different, previously untested, settings. 

We propose a new approach to modify standard neural network architectures, called Relation Based Patterns (RBP) with different variants for classification and prediction. 
Our experiments show that neural networks with the appropriate RBP structure achieve perfect classification and prediction performance on synthetic data, including mixed concrete and abstract patterns. 
RBP also improves neural network performance in experiments with real-world sequence prediction tasks. 

We discuss these finding in terms of challenges for neural network models and identify consequences from this result in terms of developing inductive biases for neural network learning.     
\end{abstract}


\section{Introduction}
Despite the impressive development of deep neural networks over recent years, there has been an increasing awareness that there are some tasks that still elude neural network learning or need unrealistic amounts of data.
Humans, on the other hand, are remarkably quick at learning and abstracting from very few examples. 
Marcus \cite{GaryMarcus1999} showed in an experiment that 7-month old infants already recognise sequences by identity rules, i.e. which elements are repeated, after just two minutes of familiarization. 
In that study %
a simple recurrent neural network model %
was also tested %
and it failed to generalise these identity rules to new data. 

In this study, we re-visit this problem and evaluate the performance of frequently used standard neural network models in learning identity rules. 
More specifically, we find that feed-forward and recurrent neural networks (RNN) and their gated variants (LSTM and GRU) in standard set-ups clearly fail to  learn general identity rules presented as classification and prediction tasks. 

We tackle this problem by proposing \emph{Relation Based Patterns} (RBP), which model identity relationships explicitly as extensions to neural networks for classification and prediction. 
We show experimentally that on synthetic data the networks with suitable RBP structures learn the relevant rules and generalise with perfect classification and prediction. 
We also show that this perfect performance extends to mixed rule-based and concrete patterns, and that RBP improves prediction on real-world language and music data. 

Identity rules are clearly in the hypothesis space of the neural networks, but the networks fail to learn them by gradient descent.
We identify that both the comparison of related input neurons and of input tokens needs to be predefined in the network to learn general rules from data. 
The RBP structures introduce this inductive bias in the neural networks and thus enable the learning of identity rules by standard neural networks.  

Our contributions in this paper are specifically: 
\begin{itemize}
	\item we evaluate several common NN architectures: feed-forward networks, RNN, GRU, and LSTM, in novel settings, and find that they fail to learn general identity rules;
	\item we identify reasons that prevent the learning process from being successful in this context;
	\item we propose the Relation Based Patterns, a new method to enable the learning of identity rules within the regular network structure;
	\item we show in experiments that identity rules can be learnt with RBP structure on artificial data, including mixed rule-based and concrete patterns, and that they improve performance in real-world prediction tasks;
\end{itemize}

The remainder of this paper is structured as follows. 
Section~2 introduces related work on modelling identity rules. 
Section~3 presents results of our experiments with standard neural network architectures. 
Section~4 presents our RBP model and its different variants. 
Section~5 presents the results of experiments using RBP structures. 
Section~6 addresses the application of RBP to mixed patterns and real data. 
Section~7 discusses the  implications of the presented experimental results and Section~8 concludes this paper.

\section{Related work}
Our task is the learning of rules from sequential data. 
This is often seen as grammar learning, on which there have been many studies in psychology.
\cite{Gordon} made an early contribution on implicit learning and generalisation.
Subsequently, \cite{Knowlton1,Knowlton2} studied specifically the knowledge acquired during artificial grammar learning tasks. ]

The specific problem we are addressing in this study is the recognition of abstract patterns that are defined by the identity relation between tokens in a sequence. 
In the well-known experiments by \cite{GaryMarcus1999}, infants were exposed to sequences of one of the forms \emph{ABA} or \emph{ABB}, e.g. `la di la' or `la di di', for a few minutes in the familiarisation phase. 

In the test phase the infants were exposed to sequences with a different vocabulary (e.g. `ba tu ba' and `ba tu tu') and they 
showed significantly different behaviour depending on whether the sequences exhibited the form they were familiarised with or not.

This type of pattern only depends on the equality between elements of the sequence and after successful learning it should be recognisable independently of the vocabulary used. 
However, \cite{GaryMarcus1999} also showed that simple recurrent Elman networks were not able to perform this learning task.
This finding sparked an exchange about whether human speech acquisition is based on rules or statistics and the proposal of several neural networks models that claimed to match the experimental results.
\cite{seidenberg1999infants} and \cite{Elman1999} proposed a solution based on a distributed representation of the input and on pre-training where the network is first trained to recognise repeated items in a sequence. 
The network is subsequently trained on classifying ABA vs ABB patterns. 
Only \cite{Elman1999} reports specific results and has only 4 test data points, but 100\% accuracy. 
However, \cite{vilcu2001generalization} reported that they could not recreate these results.

\cite{Shultz_Bale_2001} and \cite{Altmann_Dines} suggested solutions which are based on modified network architectures and training methods. 
\cite{vilcu2005two} could not replicate the results by \cite{Altmann_Dines} and found that the models by \cite{Shultz_Bale_2001} do not generalise. 
The claims by \cite{vilcu2005two} were again contested by \cite{shultz2005generalization}. A number of other methods were suggested that used specifically designed network architectures, data representations, and training methods, such as \cite{shastri1999spatiotemporal,gasser1999babies,dominey2000neural,christiansen2000connectionist}. 
More recent work by \cite{Alhama} suggests that prior experience or pre-defined context representation (``pre-training'' or ``pre-wiring'') is necessary for the network to learn general identity rules when using echo state networks. 
While these works are interesting and relevant, they do not answer our question whether more commonly used network architectures can learn general identity rules. 

The discussion of this problem is part of a wider debate on the systematicity of language learning models, which  started in the 1980s and 1990s \cite{fodor1988connectionism,hadley1994systematicity}. 
This debate, like the more specific one on identity rules, has been characterised by claims and counter-claims \cite{smolensky1987constituent,fodor1990connectionism,chalmers1990fodor,niklasson1994systematicity,christiansen1994generalization, hadley1994systematicity2}, which, as stated by \cite{frank2014getting}, often suffer from a lack of empirical grounding. 
Very recently, the work in \cite{lake2018generalization} has defined a test of systematicity in a framework of translation, applied it to standard \emph{seq2seq} neural network models \cite{sutskever2014sequence}. 
They found that generalisation occurs in this setting, but it depends largely on the amount and type of data shown, and does not exhibit the extraction and systematic application of rules in the way a human learner would.  

In most of the studies above, the evaluation has mostly been conducted 
by testing whether the output of the network shows a statistically significant difference between inputs that conform to a trained abstract pattern and those that do not. 
From a machine learning perspective, this criterion is not satisfactory as we, like \cite{lake2018generalization}, would expect that an identity rule should always be applied correctly once if it has been learned from examples, at least in cases of noise-free synthetic data. 
We are therefore interested in the question whether and how this general rule learning can be achieved with common neural network types for sequence classification. 

This question also relates to recent discussions sparked by \cite{GaryMarcus2018} about deep neural networks' need for very large amounts of training data, lack of robustness and lack of transparency as also expressed, e.g., by \cite{sabour2017dynamic,kansky2017schema,szegedy2013intriguing}. 
We surmise that these issues relate to the lack of generalisation beyond the space covered by the input data, i.e. extrapolation, which is generally seen as requiring an inductive bias in the learning system, but there is no general agreement about the nature or implementation of inductive biases for neural networks, e.g. \cite{feinman2018learning,barrett2018measuring}. 
In recent years, there was a trend to remove human designed features from neural networks, and leave everything to be learned from the data \cite{bengio2013representation}. 
We follow here the inverse approach, to add a designed internal representation, as 
we 
find that for the given problem standard neural network methods consistently fail to learn %
any suitable internal representation from the data. 

\section{Experiment 1: standard neural networks}\label{sec:exp1}

We test different network architectures to evaluate if and to what extent recurrent and feed-forward neural networks can learn and generalise abstract patterns based on identity rules. 

\subsection{Supervised learning of identity rules}

The problem in the experiment by \cite{GaryMarcus1999} is an unsupervised learning task, as the infants in the experiments were not given instructions or incentives. 
However, most common neural network architectures are designed for supervised learning and there are also natural formulations of abstract pattern recognition as supervised learning task in the form of classification or prediction. 

In our case, abstract patterns are defined by identity relations. 
Expressed in logic, they can be described using the binary equality predicate $eq(\cdot,\cdot)$.
For a sequence of three tokens $\alpha,\beta,\gamma$ the rule-based patterns $ABA$ and $ABB$ can be described by the following rules:
\begin{align}
ABA&: \neg eq(\alpha,\beta) \land eq(\alpha,\gamma)
\\
ABB&: \neg eq(\alpha,\beta) \land eq(\beta,\gamma).
\end{align}
These rules are independent of the actual values of $\alpha,\beta, \text{and} \;  \gamma$ and also called abstract patterns. 
Concrete patterns, on the other hand, are defined in terms of values of from a vocabulary %
$a,b,c, ...$ . 
E.g., sequences $a**$, i.e. beginning with `$a$', or $*bc$, ending with `$bc$', can be formulated in logic as follows:
\begin{align}
\textit{a**}&: \alpha = \mlq{}a\mrq \\
\textit{*bc}&: \beta = \mlq{}b\mrq \land \gamma = \mlq{}c\mrq.
\end{align}
For the remainder of this article we use the informal notations $ABA$ and $a**$ as far as they are unambiguous in their context.

For classification, the task is to assign a sequence to a class, i.e. \emph{ABA} or \emph{ABB}, after learning from labelled examples. 
For prediction, the task is to predict the next token given a sequence of two tokens after exposure to sequences of one of the classes (e.g. only ABA, or ABB respectively). 
These tasks are suitable for the most commonly used neural network architectures. 

\subsection{Experimental set-up}\label{sec:exp-set-up}
\paragraph{Network set-up}
We use the Feed-forward Neural Network (FFNN) (also called Multi-layer Perceptron) \cite{rumelhart1985learning}, the Simple Recurrent Neural Network (RNN, also called Elman network \cite{elman1990finding}), the Gated Recurrent Unit (GRU) network \cite{cho2014learning}, and the Long Short Term Memory  (LSTM) network  \cite{hochreiter1997long}.
For Prediction we only use the RNN and its gated variants GRU and LSTM.

The input to the networks is a one-hot encoded vector representing each token with $n$ neurons, where $n$ is the size of the vocabulary.   
In the case of the FFNN, we encode the whole sequence of 3 tokens as a vector of size $3n$. 
For the recurrent models, we present the tokens sequentially, each as a $n$-dimensional vector. 
We set the number of neurons in each hidden layer to (10, 20, 30, 40, 50), using 1 or 2 hidden layers. 
We use Rectified Linear Units (ReLUs) for the hidden layers in all networks. 
The output layer uses the softmax activation function. 
The number of output units is 2 for  classification and the size of the vocabulary for prediction.
We train with the Adam optimisation method \cite{kingma2014adam}, using initial learning rates of $0.01, 0.1, 0.2, 0.4$, and train with the synthetic datasets in one batch.
We use regularisation with Dropout rates of 0.1, 0,2, 0.4 and set the number of epochs to 10, within which all trainings converged. 

We conduct a full grid search over all  hyperparameters using four-fold cross-validation to optimise the hyperparameters and determine test results. 
We run a total of 10 simulations for each evaluation and average the results. 
All experiments have been programmed in PyTorch and the code is publicly available.\footnote{\url{https://github.com/radhamanisha1/RBP-architecture}}

\paragraph{Datasets}
For performing the rule learning experiments, we artificially generate data in the form of triples for each of the experiments. 
We consider our sample 
vocabulary as $a ... l$ (12 letters) for both prediction and classification tasks. 
We generate triples in all five abstract patterns: AAA, AAB, ABA, ABB, and ABC for the experiments. 
The sequences are then divided differently for the different cases of classification. 
For all the experiments we use separate train, validation, and test sets with 50\%, 25\%, and 25\% of the data, respectively. 
All sampling (train/test/validation split, downsampling) is done per simulation. 

\subsection{Classification} 
First we test three different classification tasks as listed below. 
We use half the vocabulary for training and the other half for testing and validation (randomly sampled). 
We divide the sequences into two classes as follows, always maintaining an equal size of both classes:
\begin{itemize}
\item[1)] ABA/ABB vs other: In task a) class one contains only pattern ABA while the other contains all other possible patterns (AAA, AAB, ABB, ABC) downsampled per pattern for class balance. 
The task is to detect whether $eq(\alpha,\gamma) \land \neg eq(\alpha,\beta)$ is true or false. 
Analogously, the task in b) ABB vs other is to detect $eq(\beta,\gamma) \land \neg eq(\alpha,\beta)$. 
This case corresponds to the experiment in \cite{GaryMarcus1999}, where only one rule-based pattern type is used for familiarisation. 

\item[2)] ABA vs ABB: This task is like task 1 above, but only pattern ABB occurs in the second class, so that this task has less variance in the second class. 
We expected this task to be easier to learn because two equality predicates $eq(\alpha,\gamma), eq(\beta,\gamma)$ change their values between the classes and are each sufficient to indicate the class. 

\item[3)] ABC vs other: In this case, class one (ABC) has no pair of equal tokens, while the $other$ class has at least one of $eq(\alpha,\beta), eq(\alpha,\gamma), eq(\beta,\gamma)$ as $true$, i.e. detecting equalities without localising them is sufficient for correct classification. 
\end{itemize}
In our experiments, the training converged quickly in all cases and the classification accuracy on the training data was 100\%. 
The results on the test set are shown in Table~\ref{tab:std-nn-class}.
In all cases the baseline, corresponding to random guessing is 50\%. 
This baseline is only exceeded for task 1) by the RNNs and their gated variants, and even then the accuracy is far from perfect at 55\%. 
\begin{table}[ht]
    \centering
    \begin{tabular}{|l|r|r|r|r|} \hline
    Classification task & FFNN & RNN & GRU & LSTM \\ \hline
    1a) ABA/other &  50\%    &   55\%  &   55\% &   55\%  \\
    1b) ABB/other  &  50\%    &   55\%  &   55\% &   55\% \\
    2) ABA/ABB &  50\%    &  50\%   &  50\% &  50\% \\
    3) ABC/other & 50\%    &  50\%   &  50\% &  50\% \\\hline
    \end{tabular}
    \caption{Three classification tasks based on abstract patterns over 10 simulations.
    The numbers show test set accuracy after a grid search and cross validation as described in section \ref{sec:exp-set-up}. 
    All values are rounded to the next percentage point.}
    \label{tab:std-nn-class}
\end{table}

\subsection{Prediction} 
We performed prediction experiments on two tasks. In task 1) we train and test on ABA patterns and in task 2) on ABB. 
Training and test/validation set use different vocabularies. 
The training converged quickly in less than 10 epochs, and after training the classification accuracy on the training set is 100\%.

The results on the test set are shown in Table~\ref{tab:std-nn-pred}.
The baseline is $~8.3 \ldots \%$ 
as we have a vocabulary size of 12. We use again half the vocabulary (6 values) for training and half for validation/testing.
The results show that the tested networks fail completely to make correct predictions. 
They perform below the baseline at 0\% accuracy, which is mostly because they predict only tokens that appear in the training set but not in the test set. 

\begin{table}[ht]
\centering
\begin{tabular}{|l|r|r|r|}
\hline
Prediction task & RNN  & GRU & LSTM\\
\hline
1) ABA & 0\% & 0\% & 0\% \\
2) ABB & 0\% & 0\% & 0\%  \\
\hline
\end{tabular}
\caption{Prediction results for two different abstract patterns. 
The numbers show test set prediction accuracy after a grid search and cross validation as described in section \ref{sec:exp-set-up}. 
}\label{tab:std-nn-pred}
\end{table}

\subsection{Discussion}
The results show clearly that FFNNs, RNNs, GRUs and LSTMs do not learn general abstract patterns based on identity rules. 
This agrees with the previously reported experiments by \cite{GaryMarcus1999}. 
However, since there was some conflicting evidence in the literature, the clarity of the outcome was not expected. 

\paragraph{Questions raised}
This result raises the question of why these neural networks do not learn to generalise abstract patterns from data.
There are two aspects worth considering for an explanation: the capacity of the network and the necessary information for the network to solve the problem. 

Regarding the capacity: the solution to the task is in the hypothesis space of the neural networks, since proofs exist of universal approximation properties for feed-forward networks with unbounded activation functions \cite{leshno1993multilayer} and of Turing-completeness for recurrent networks \cite{Siegelmann}. 
We will present a constructive solution below, putting that result into practice, with a design of network instances that solve the problem. 

The relevant question, as has been pointed out by \cite{Alhama}, is therefore why learning with backpropagation does not lead to effective generalisation here. 
There are three different steps that are necessary to detect identity rules: a comparison of input neurons, a comparison of tokens, represented by multiple neurons, and a mapping of comparison results to classes or predictions.

\paragraph{Vocabulary hypothesis}
A possible reason for the failure of the networks to generalise what we call the vocabulary hypothesis. 
It is based on the separated vocabulary in one-hot encoded representation. 
This leads to some input neurons only being activated in the training set and some only in the validation and test sets. 

In order to learn suitable weights for an input comparison, there would have to be a suitable gradient of the weights of the outgoing connections from these inputs.
If parts of the vocabulary do not appear in the training data, i.e. the activation of the corresponding input neurons is always zero during training, the weights of their outgoing connections will not be adapted. 
We therefore expect that the separation of the vocabulary prevents generalisation from the training to the test set as the weights going out from neurons that are used during testing have not been adapted by the gradient descent.
Based on this consideration we conducted another experiment with a shared vocabulary.

This experiment is called ABA-BAB vs other. 
We again represent our vocabulary as $a...l$ (12 letters) for this task with train/validation/test split as $50\%/25\%/25\%$.
Now we use the same vocabulary for training, validation, and testing, but we separate different sequences of the form ABA that use the same tokens between the training and validation/test sets. 
E.g., if the $ded$ is in the test set, then $ede$ is in the training or validation set, so that there is no overlap in terms of actual sequences. 
Like in classification experiment 1), training converged quickly and resulted in perfect classification performance on the training set.

\begin{table}[ht]
    \centering
    \begin{tabular}{|l|c|c|c|c|} \hline
    Classification task & FFNN & RNN & GRU & LSTM\\ \hline
    ABA-BAB vs other & 50\%    &  50\%   & 50\% & 50\%\\ \hline
    \end{tabular}
    \caption{Classification results on test sets with the same vocabulary used in test, validation and training set.}
    \label{tab:std-nn-shared}
\end{table}

The results on the test set presented in Table~\ref{tab:std-nn-shared} show performance at the baseline with no evidence of generalisation. 
This shows that activating all inputs by using a shared vocabulary is not sufficient to enable generalisation in the learning process. 

\paragraph{Other explanations}
A second potential problem is which neurons should be compared. 
The FFNN has no prior information about neurons belonging to the same or different tokens or about which input neurons correspond to the same token values. 
In the RNN, one token is presented per time step, so that a comparison between the previous hidden state and the current input is possible as the same neurons are activated. 
However, with a full set of connections between the previous hidden layer and the current, there is no reason that relations between the same neurons at different time steps would be processed differently from different neurons.

On the other hand, if we had a representation that includes the information of which tokens are identical or different, then we would have all the information we need for a mapping, as these are the relations in which our defining rules are formulated (e.g. ABA is defined as $eq(\alpha,\gamma) \land \neg eq(\alpha,\beta)$). 
This idea has led to the Relation Based Pattern (RBP) model that we introduce in the next section and then evaluate with respect to its effect on both abstract and concrete pattern learning.

\section{Design of Relation Based Pattern models}

To address the inability of neural networks to generalise rules in neural network learning, we developed the Relation Based Pattern (RBP) model as a constructive solution, where the comparisons between input neurons and between tokens and the mappings to outputs are added as a predefined structure to the network. 
The purpose of this structure is to enable standard neural networks to learn abstract patterns based on the identity rules over tokens while retaining other learning abilities.  

In the RBP model there are two major steps. 
The first step is defining comparison units for detecting repetitions, called DR units, and the second step is adding the DR units to the neural network. 

\subsection{Comparison units}

\begin{figure}[htp] 
    \centering{
        \includegraphics[width=0.5\textwidth]{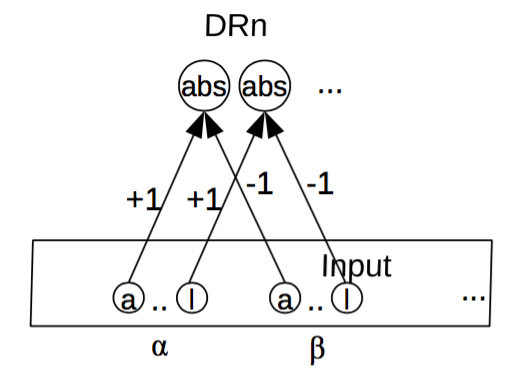}
        }%
	\caption{$DR_n$ units comparing related inputs with an absolute of difference activation function. 
		In one-hot encoding there are $k  DR_n$ units for every pair of input tokens, where $k$ is the vocabulary size.}
	\label{fig:drunit1}
\end{figure}

\paragraph{Comparing neurons}
We assume, as before, that input is a one-hot encoded vector of the current token along with the $n-1$ previous vectors for a given context length \textit{n} (in this study context length 3 for classification and 2 for prediction). 
We use comparison units, called DR units (differentiator-rectifier). 
As the name suggests, they apply a full wave rectification to the difference between two inputs: $f(x,y) = |x-y|$. 
The first level of $DR$ units are $DR_n$ units that are applied to every pair of corresponding input neurons (representing the same value) within a token representation, as shown in Figure~\ref{fig:drunit1}.

\begin{figure*}
\centerline{
    \hfill
    \subfloat[The $DR_p$ and $DR_n$ units that are used in the RBP1 and RBP2 structures with $3 \times k$ $DR_n$ and 3 $DR_p$ units for a vocabulary size $k$ and sequence length~3.]{%
        \includegraphics[width=0.35\textwidth]{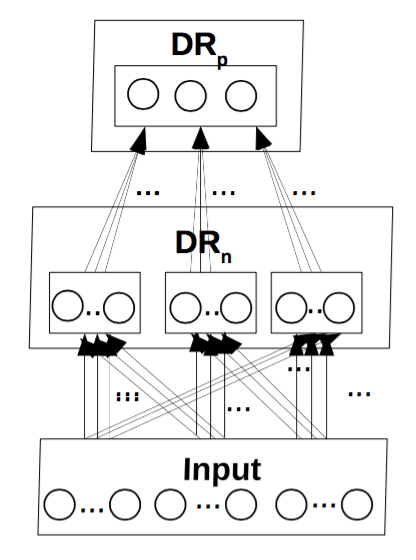}%
        \label{fig:dr_in_plus}%
        }%
    \hfill
    \subfloat[The $DR_{out}$ structure for detecting repetitions between input and target. The $DR_{p}out$ values are calculated at training time and a model is trained to predict them conditional on $DR_pin$ (see Figure~\ref{fig:rbp3}).]{%
        \includegraphics[width=0.40\textwidth]{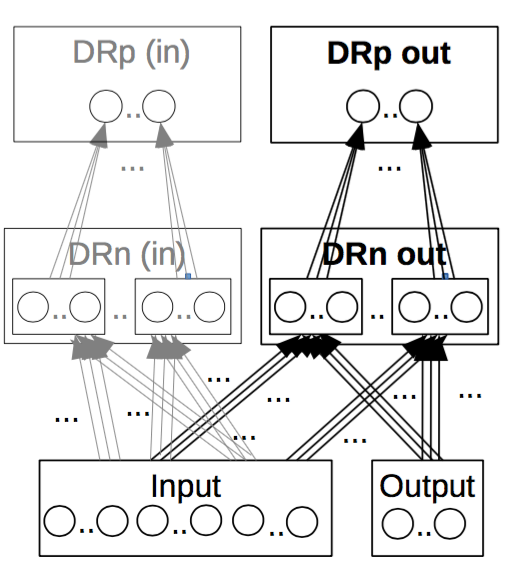} 
        \label{fig:dr_out}
        }
    \hfill
    }

  \caption{$DR_n$ and $DR_p$ units for inputs (all RBP) and outputs (RBP3).}
 \label{fig:dr_over}
\end{figure*}

\paragraph{Comparing tokens}
The next level of $DR$ units are the $DR_p$ units that sum the activations of the $DR_n$ values that belong to one pair of tokens.
Based on the sequence length $n$ and vocabulary size $a$ we create $k = a \times n(n-1)/2$ $DR_{n}$ units for all the possible pairs of tokens and  i.e. in our classification example, we have a sequence of 3 tokens and a vocabulary size of 12, i.e. $12 \times 3(3-1)/2 = 36 \times 3$ $DR_n$ units. 
All the $DR_n$ units for a pair of tokens are then summed in a $DR_p$ unit using connections with a fixed weight of +1. 
E.g. we have $5 \times (5-1)/2 = 10$ $DR_p$ units for a context of length 5.  
Figure \ref{fig:dr_in_plus} shows the network structure with $DR_n$ and $DR_p$ units.

For the prediction case, we also use the same approach to represent the difference between each input token and the next token (i.e., the target network output). 
We create $n$ $DR_p out$ units that calculate the difference between each input in the given context and the next token. 
There are $k \times n$ $DR_n {out}$ units that compare the corresponding neurons for each pair of input/output tokens, in the same way as for the pairs of input tokens.
The overall network structure is shown in Figure~\ref{fig:dr_out}.

\subsection{Neural network integration}

\begin{figure}[htp] 
    \centering{
        \includegraphics[width=0.5\textwidth]{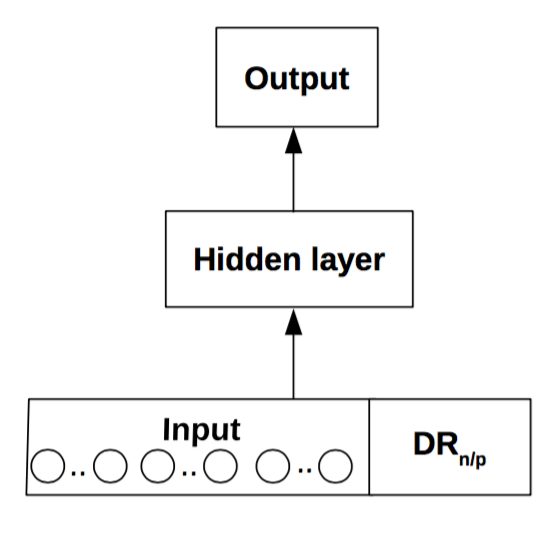}
        }%
	\caption{Overview of the RBP1n/RBP1p structure.}\label{fig:rbp1} 
\end{figure}

We combine the DR units ($DR_{n}$ and $DR_p$) with the neural network models in early, mid and late fusion approaches we call RBP1, RBP2 and RBP3, as outlined below. 
The weights that connect $DR_n$ units to input and output, and the $DR_n$ to $DR_p$ units and the offset layer are fixed, all other weights that appear in the following models are trainable with backpropagation.

\paragraph{Early Fusion (RBP1n/p)}
In this approach, $DR_n$ or $DR_p$ units are added as additional inputs to the network, concatenated with the normal input. 
In Figure \ref{fig:rbp1}, the RBP1n/p structure is depicted.
We use early fusion in both the prediction and classification tasks.

\begin{figure*}
\centering
 \centerline{\hspace{5mm}
 	\subfloat[RBP2a]{\includegraphics[width=0.4\textwidth]
 {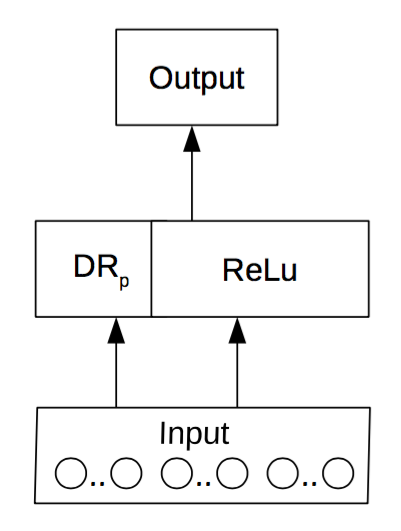}\label{fig:rbp2a}}
 \hfill
 \subfloat[RBP2b]{\includegraphics[width=0.45\textwidth]{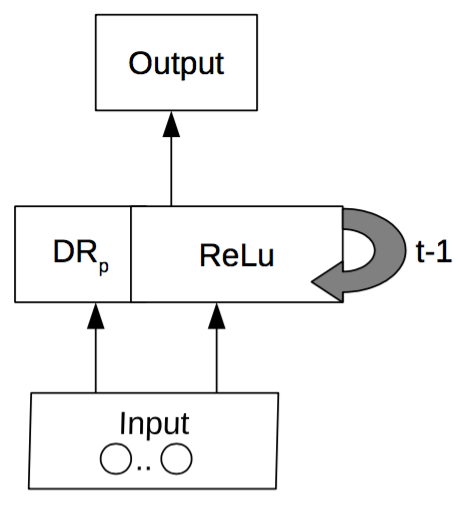}\label{fig:rbp2b}}}
 \caption{Overview of RBP2 approaches, where the $DR_p out$ units are concatenated to the hidden layer.}
 \label{fig:rbp2}
\end{figure*}

\paragraph{Mid Fusion (RBP2)}
The $DR_p$ units are added to the hidden layer. 
Figure \ref{fig:rbp2a} shows the mid fusion structure for the feed-forward network and Figure \ref{fig:rbp2b} for the recurrent network respectively. 
The RBP2 approach is used for classification and prediction tasks. 

\begin{figure*}
\centerline{\includegraphics[width=6.49cm]
 {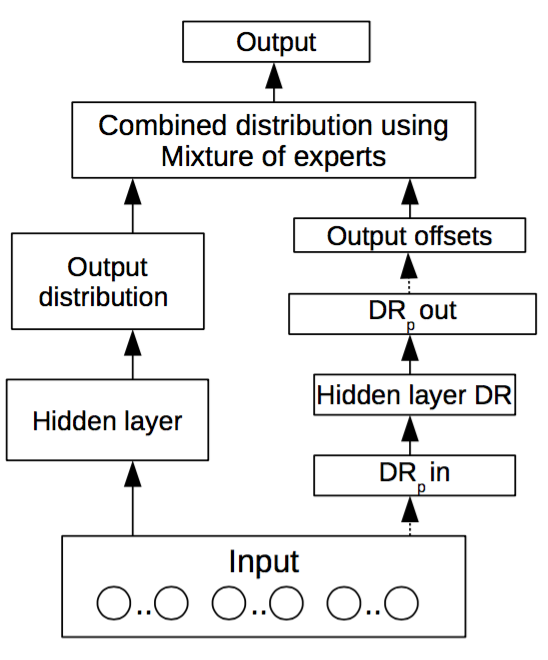}}
  \caption{Overview of the RBP3 approach. 
  The $DR_p in$ values are calculated as in RBP2. 
  From there, we use a fully connected layer to predict $DR_p out$ (trained with teacher-forcing). 
  The predicted $DR_p {out}$ values are mapped back to the vocabulary (based on the context tokens) and used as probability offsets in a mixture of experts with the standard neural network in the left part of the diagram. 
  All connections are trainable except $Input$ to $DR_p in$ and $DR_p out$ to \textit{Output offsets} (dotted arrows).}
 \label{fig:rbp3}
\end{figure*}

\paragraph{Late Fusion (RBP3)}
In this approach, we use the same structure as in RBP2 (we call it $DR_n in$ and $DR_p in$ in this context), and in addition we estimate the probability of identity relations between the input and the output, 
i.e., that the token in the current context is repeated as the next token. 
We use a structure called $DR_p out$ for this, and from there we project back to the vocabulary, to generate a probability offset for the tokens appearing in the context.

Figure \ref{fig:rbp3} gives an overview of the RBP3 late fusion scheme. 
The $DR_p in$ units detect identities between the input tokens in the current context as before. 
The $DR_p out$ units model the identities between the context and the next token, as shown in the Figure \ref{fig:rbp2b}, where repetition is encoded as 1, and a non-repeated token as a $-1$. 
During training we use teacher-forcing, i.e., we set the values of the $DR_p out$ units to the true values. 
We use a feed-forward neural network with one hidden layer to learn a mapping from the $DR_{in}$ to the $DR_{out}$. 
This gives us an estimate of the $DR_{out}$ units given the $DR_{in}$ units. 
The $DR_{out}$ values are then normalised subtracting the mean, and then mapped back to the output space (the one-hot vocabulary representation), using a zero value for the output values that don't appear in the input. 
These output offsets are then combined in a weighted sum (mixture of experts) with the output distribution estimated by the standard normal network (on the left side in Figure \ref{fig:rbp3}).
The weights in the mixture are trainable.
The outputs from the combined distribution of mixture of experts are clipped
between [0,1] and renormalised. 
The final output distribution is a softmax layer providing the probability distribution over the vocabulary for the next token.

\section{Experiment 2: neural networks with RBP structures}

In the following we repeat the experiments from section \ref{sec:exp1} but also test
networks with added RBP structures. For convenience we repeat the previous results in the tables in this section.  

\subsection{Classification experiments}
This experiment is analogous to the first classification experiment. 
In the case of the feed-forward network, RBP2(a) was used in the mid fusion approach and for recurrent network, RBP2(b) was used. 
We trained again for 10 epochs and all networks converged to perfect classification on the training set. 
Table \ref{tab:rbp-cl-res} provides the overall test accuracy for the three approaches. 

\begin{table}[ht]
\begin{center}
\centering
\begin{tabular}{|l|l|r|r|r|r|}
\hline
Task &  RBP & FFNN  & RNN & GRU & LSTM\\ \hline

 \multirow{4}{*}{1a) ABA vs other} & -  & 50\% & 55\%  & 55\% & 55\%\\
 & RBP1n & 50\% & 55\% & 55\% & 55\% \\
 & RBP1p & 65\% & 70\% & 70\% & 70\% \\
 & RBP2 & 100\% & 100\% & 100\% & 100\%\\ \hline
 
\multirow{4}{*}{1b) ABB vs other}& - & 50\% & 55\%  & 55\% & 55\%\\
    & RBP1n & 50\% & 55\% & 55\% & 55\% \\ 
    & RBP1p & 65\% & 70\% & 70\% & 70\% \\
    & RBP2 & 100\% & 100\% & 100\% & 100\%\\ \hline
   
\multirow{4}{*}{2) ABA vs ABB}& - & 50\% & 50\% & 50\% & 50\%\\
    & RBP1n & 50\% & 60\% & 65\% & 65\%\\ 
    & RBP1p & 75\% & 75\% & 75\% & 75\%\\
    & RBP2 & 100\% & 100\% & 100\% & 100\%\\ \hline

\multirow{4}{*}{3) ABC vs other}& - & 50\% & 50\%  & 50\% & 50\%\\
    & RBP1n & 55\% & 65\%  & 65\% & 65\%\\ 
    & RBP1p & 55\% & 70\% & 70\%  & 70\%\\
    & RBP2 & 100\% & 100\% & 100\% & 100\%\\ \hline
    
\multirow{4}{*}{4) ABA-BAB vs other}& - & 50\% & 50\% & 50\% & 50\%\\
    & RBP1n & 55\% & 72\% & 75\% & 75\%\\ 
    & RBP1p & 69\% & 74\% & 75\% & 76\%\\
    & RBP2 & 100\% & 100\% & 100\% & 100\%\\ \hline
\end{tabular}
\end{center}
\caption{Classification experiments with RBP: test accuracy for the different models and tasks, as explained above. 
Results with `-' in the RBP column are the same as in section \ref{sec:exp1} and shown here again for comparison.}
\label{tab:rbp-cl-res}
\end{table}

The results with RBP1n models already show some improvement over the baseline in most configurations, but the result are only slightly above the standard networks, with RNNs, GRUs and LSTMs benefiting more than FFNN. 
This supports our hypothesis that learning to compare corresponding input neurons is a challenging task for neural networks. 
However, the results show that providing that comparison is not sufficient for learning identity rules. 

RBP1p structures also aggregate all the $l$ $DR_n$ neurons that belong to a pair of input tokens. 
The results show that providing that information leads to improved accuracy and provide evidence that this aggregation is another necessary step that the networks do not learn reliably from the data.

The RBP2 models enable the neural networks to make predictions and classifications that generalise according to identity rules that it learns from data.
The RBP2 leads to perfect classification for all network types tested. 
This confirms the design consideration that comparing pairs of tokens provides the relevant information in the form required  for classification, as the classes are defined by $equals$ relations, so that the activations of the DRp units are directly correlated with the class labels. 

A surprising result is the big difference between the generalisation using the RBP1p and the RBP2 structures. 
They both provide the same information, only in different layers of the network, but RBP1p only reaches at most 75\% with a 50\% baseline. 
We hypothesize that the additional expressive power provided by the non-linearities in the hidden layer here provides effective learning. 
This effect deserves further investigation. 

\subsection{Prediction experiments}
Here we performed two experiments separately on ABA and ABB patterns as in experiment 1 on prediction. 
The tasks are the same as previously and we trained again for 10 epochs after which all networks had converged to perfect prediction accuracy on the training data.  
Table \ref{tab:pred-res}, summarises the accuracy for RNN, GRU and LSTM without RBP, and with RBP1n, 1p, 2, and 3. 

\begin{table}[ht]
\centering
\begin{tabular}{|c|c|r|r|r|}
\hline
Pattern & RBP & RNN  & GRU & LSTM\\
\hline
\multirow{5}{*}{1) ABA} & - & 0\% & 0\% & 0\% \\
 & RBP1n & 0\% & 0\% & 16\%  \\
 & RBP1p &  0\% & 0\% & 18\%  \\ 
& RBP2 & 0\% & 0\% & 20\%  \\
& RBP3 & 100\% & 100\% & 100\%  \\
\hline
\multirow{5}{*}{2) ABB} & - &  0\%  & 0\% &  0\% \\
 & RBP1n & 0\% & 0\% & 17\%  \\
 & RBP1p &   0\%  &  0\% &  20\% \\
& RBP2 & 0\% & 0\% & 22\% \\
& RBP3 & 100\% & 100\% & 100\% \\
\hline
\end{tabular}
\caption{Test set accuracy in prediction experiments for patterns ABA and ABB. 
As before, results are averaged over 10 simulations and rounded to the nearest decimal point. 
Results with ‘-’ in the RBP column are the same as in section 3 and shown here again for comparison.} 
\label{tab:pred-res}

\end{table}

Overall, we observe that only the LSTM benefits from RPB1n, RBP1p, and RBP2 structures, all other networks can apparently not make use of the information provided. 
The RBP3 model, on the other hand, leads to perfect classification on our synthetic dataset. 

Our interpretation is, that standard recurrent networks do not learn the more complex mapping that prediction requires, as not only recognition of a pattern but also selecting a prediction on the basis of that pattern is required. 
The somewhat better results of the LSTM networks are interesting. 
In the RBP3 model, the mapping between the identity patterns and back to the vocabulary adds considerable prior structure to the model and it is very effective in achieving the generalisation of rule-based patterns. 

\section{Experiment 3: mixed tasks and real data}

The results presented here were all obtained with synthetic data where classification was exclusively on rule-based abstract patterns. 
This raises the question whether the RBP will impede recognition of concrete patterns in a mixed situation. 
Furthermore, we would like to know whether RBP is effective with real data where the abstract and concrete patterns may interact. 

\subsection{Mixed abstract and concrete patterns}

We conducted an experiment where the classes were defined by combinations of abstract and concrete patterns. 
Specifically we defined 4 classes based on the abstract patterns $ABA$ and $ABB$ combined with the concrete patterns $a**$ and $b**$.  
E.g., the class $ABA,a**$ can be expressed logically as 
\begin{equation}
eq(\alpha,\gamma) \land \neg eq(\alpha,\beta) \land \alpha = `a\mrq. 
\end{equation} 
We use a vocabulary of 18 characters, out of which 12 are used for training and 6 are used for validation/testing in addition to `a' and `b', which need to appear in all sets because of the definition of the concrete patterns.
For class 1/3 and class 2/4, abstract patterns ABA and ABB are used respectively. 
Class 1/2 and 3/4 start with tokens `a' and `b' respectively. 
The train, validation and test split is 50\%, 25\%, and 25\% respectively. 
We trained the network for 10 epochs, leading to perfect classification on the training set.
A total of 10 simulations has been performed.
We test a feed forward and a recurrent neural network without and with RBP1p and RBP2. 
The results are shown in Table~\ref{tab:mixed}.
\begin{table}[ht]
    \centering
    \begin{tabular}{|l|r|r|}\hline
        RBP &  FFNN & RNN \\ \hline
        - & 23\%  & 42\% \\
        RBP1p & 49\%  & 57\% \\
        RBP2 & 100\%  & 100\% \\ \hline
    \end{tabular}
    \caption{Test set accuracy for mixed abstract/concrete pattern classification.}
    \label{tab:mixed}
\end{table}

As in the previous experiments, networks without RBP fail to generalise the abstract patterns. 
The results for RBP1p and RBP2 show, that the ability to learn and recognise the concrete patterns is not impeded by adding the RBP structures.  

\subsection{Language models with RBP}
In order to test the capability of networks with RBP structure, we use them in two language modelling tasks. 
One is to predict characters in English text, and one is to predict the pitch of the next note in folk song melodies.
We selected both tasks because of the prevalence of repetitions in the data, as notes in music and characters in English tend to be repeated more than words. 
Our RBP structures are designed to model identity-rules and we therefore expect them to be more effective on tasks with more repetitions. 

\paragraph{Character prediction}
We conducted a character prediction experiment on a 
subset of the Gutenberg electronic book collection\footnote{https://www.gutenberg.org/}, consisting of text with the dataset size of 42252 words. 
We used 2 hidden layers with 50 neurons each. 
In the RBP2 model, the DRp units were concatenated with the first hidden layer. 
The learning rate is set to 0.01 and the network training converged after 30 epochs. 
Each character is predicted without and with the RBP variants using a context size of 5.
The prediction results are summarized in Table \ref{tab:charrnn}.

\newcommand{\bftab}{\fontseries{b}\selectfont}

\begin{table}[ht]
    \centering
\begin{tabular}{|l|r|r|r|}
\hline
RBP        & RNN    & GRU    & LSTM   \\ \hline
-  & 3.8281 & 3.8251 & 3.8211 \\
RBP1p & 4.4383 & 4.4368 & 4.4321 \\
RBP2   & 3.7512 & 3.7463 & 3.7448 \\
RBP3  & \bftab{3.4076} & \bftab{3.4034} & \bftab{3.4012} \\ \hline
\end{tabular}
\caption{Character prediction task. 
The numbers show the average cross entropy loss per character on the test set (lower is better, best values are set in bold), without and with RBP structures using context length 5.}
\label{tab:charrnn}
\end{table}

\paragraph{Pitch prediction}
In another experiment we applied RBP to pitch prediction in melodies \cite{icml} taken from the Essen Folk Song Collection \cite{Schaffrath1995}. 
We performed a grid search for each context length for hyper parameter tuning, with [10,30,50,100] as the size of the hidden layer and [30,50] epochs with learning rate set to 0.01 with one hidden layer.
The results for context length 5 are summarized in Table~\ref{tab:icml}. RBP improved the network performance for RNN, GRU, and LSTM.  
Overall, LSTM with late fusion produces the best result and also improves over the best reported performance in pitch prediction with a long-term model on this dataset with a cross-entropy of 2.547, which was achieved with a feature discovery approach by \cite{Martin}.

\begin{table}[ht]
    \centering
\begin{tabular}{|l|r|r|r|}
\hline
RBP & RNN & GRU & LSTM    \\ \hline
- & 2.6994 & 2.5702 & 2.5589 \\
RBP1p & 2.6992 & 2.5714 & 2.5584 \\ 
RBP2 & 2.6837 & 2.5623 & 2.5483 \\ 
RBP3 & \bftab{2.6588} & \bftab{2.5549} & \bftab{2.5242} \\ \hline
\end{tabular}
\caption{Pitch prediction task on the Essen Folk Song Collection. 
The numbers show the average cross entropy per note (lower is better, best values are set in bold), without and with RBP using context length 5.}
\label{tab:icml}
\end{table}

\paragraph{Results}
In both character and pitch prediction, the addition of RBP3 structures improves the overall results consistently. 
RBP1n leads to a deterioration in character prediction and to inconsistent effect on pitch prediction, while RBP2 leads to a slight but consistent improvement in both tasks. 
This provides further evidence that the RBP structure enables the learning of relevant patterns in the data. 

\section{Discussion}

\subsection{Standard neural networks}
The results of the experiments described above confirm the results 
of \cite{GaryMarcus1999} and others that standard recurrent (and feed-forward) neural networks do not learn generalisable identity rules. 
From the tested models and settings of the task we can see that the lack of activation of input neurons impedes learning, but avoiding this lack is not sufficient. 
The task assumes that the identity of input tokens is easy to recognise, classify and base predictions on, but the models we tested do not learn to generalise in this way. 
These results confirm our view that in order to generalise it is necessary to know which input neurons are related, similarly on the next level, which comparisons of input belong to a pair of tokens so that they can be aggregated per token. 
The structure of neural networks does not provide any prior preference for inputs that are related in this way over any other combinations of inputs. 
This makes it seem plausible that the solutions by \cite{seidenberg1999infants,Elman1999,Altmann_Dines} could not be replicated by \cite{vilcu2001generalization,vilcu2005two}. 

\subsection{Constructive model with RBP}

The RBP model addresses the learning of identity rules by adding neurons and connections with fixed weights. 
From the input neurons we add connections to a DR (differentiator-rectifier) unit from each pair of corresponding input neurons within any pair of tokens (represented in one-hot encoding).
These DR units calculate the absolute of the difference of the activations of the two input neurons. 
They are followed by $DRp$ units that aggregate by taking the sum of the DR unit activations for each pair of tokens. 
The fact that the DRp units relate to the difference between each pair of neurons makes the learning task for classification much simpler, as has been confirmed by our results.
An open question in this context is why the RBP2  is so much more effective than RBP1p for classification, although the only difference is the layer in which the information is added into the network.

For prediction, we need a more complex structure, as beyond recognition of identity, also the selection of the token to predict is required, that depends on the tokens in the context and their similarity relations. 
The constructive RBP solution requires a transformation into a representation of identity relations in the input that is mapped to identities between input and output and that is mapped back to the token space by adding prediction probability to the tokens that are predicted to be identical between input and output. 
This created a complex predefined structure, but without it even the models that achieved prefect classification failed to make correct predictions with new data. 
Only the LSTM models could use the RPB1 and RBP2 information to make prediction above the baseline (22\% vs 8.3\%). 
We hypothesise that the gating structure of the LSTMs enables at least some effective mapping.
The 100\% correct predictions by all models using RBP3 shows the effectiveness of this structure.

\subsection{Applications}

Adding a bias into the network with a predefined structure such as RBP raises the question whether there is a negative effect on other learning abilities of the network and whether interactions between the abstract and concrete tasks can be learnt. 
In the mixed pattern experiment, RBP is still effective and showed no negative effect. 
In experiments with real language and music data we found that RBP3 has a positive effect on the prediction of characters in language and pitches in folk song melodies. 
The small negative effect of RBP1 on character prediction seems to indicate that there may be confounding effect where identity rules are less relevant. This effect did not appear in melody prediction, where repetition is more important.   

\subsection{Extrapolation and inductive bias}

The results in this study confirm that an inductive bias is needed for extrapolation, in the terminology of  \cite{GaryMarcus2018}, in order to generalise in some sense outside the space covered by the training data. 
This general challenge has recently attracted some attention. 
E.g., 
\cite{mitchell2018extrapolation} provided several solutions to the related problem of learning equality of numbers (in binary representation), which does not generalise from even to odd numbers as pointed out already by \cite{GaryMarcus2001}. 
As the authors point out in  \cite{mitchell2018extrapolation}, an essential question is which biases are relevant to the domain and problem. 
The identity problem addressed here is in itself fundamental to learning about relations \cite{battaglia2018relational}, as relations depend on object identity. 
This further raises the question what is needed to enable more complex concepts and rules to be learnt, such as more general logical concepts and rules.  

The identity rules also point to the lower-level problem that the natural relations of position and belonging to objects are not naturally addressed in neural networks. 
Other tasks may require different structures, relating for example to arithmetics, geometry or physics \cite{cohen2016inductive}.  
We therefore see as an important task the definition or predefined structures in neural networks, so that they create useful inductive bias, but do not prevent learning of functions that do not conform to that bias. 

\section{Conclusions}
Our experiments show that the observation by \cite{GaryMarcus1999}, that neural networks are unable to learn general identity rules, holds for standard feed-forward networks, recurrent neural  networks, and networks of GRUs and LSTMs. 
The solution we propose here, the Relation Based Patterns (RBP), introduce an additional structure with fixed weights into the network. 
Our experiments confirm that the RBP structures enable the learning of abstract patterns based on identity rules in classification and prediction as well as in mixed abstract and concrete patterns.  
We have further found that adding RBP structures improves performance in language and music prediction tasks.  

Overall, we find  that standard neural networks do not learn identity rules and that adding RBP structure creates an inductive bias which enables this extrapolation beyond training data with neural networks.  
This outcome raises the question on how to develop further inductive biases for neural networks to improve generalisation of learning on other tasks and more generally. 

\bibliographystyle{elsarticle-num}
\bibliography{research}
\end{document}